\documentclass{article}

\usepackage[preprint]{neurips_2019}

\usepackage[utf8]{inputenc} 
\usepackage[T1]{fontenc}    
\usepackage{hyperref}       
\usepackage{url}            
\usepackage{booktabs}       
\usepackage{amsfonts}       
\usepackage{nicefrac}       
\usepackage{microtype}      
\usepackage{xcolor}
\usepackage{worrall}
\usepackage{multicol}
\usepackage{wrapfig}
\usepackage{makecell}

\title{Deep Scale-spaces: Equivariance Over Scale}

\author{
  Daniel E.~Worrall\thanks{deworrall92.github.io} \\
  AMLAB, Philips Lab\\
  University of Amsterdam\\
  \texttt{d.e.worrall@uva.nl} 
  \And
  Max Welling \\
  AMLAB, Philips Lab\\
  University of Amsterdam\\
  \texttt{m.welling@uva.nl}
}

\begin{document}

\maketitle

\begin{abstract}
We introduce deep scale-spaces (DSS), a generalization of convolutional neural networks, exploiting the scale symmetry structure of conventional image recognition tasks. Put plainly, the class of an image is invariant to the scale at which it is viewed. We construct scale equivariant cross-correlations based on a principled extension of convolutions, grounded in the theory of scale-spaces and semigroups. As a very basic operation, these cross-correlations can be used in almost any modern deep learning architecture in a plug-and-play manner. We demonstrate our networks on the Patch Camelyon and Cityscapes datasets, to prove their utility and perform introspective studies to further understand their properties.
\end{abstract}

\section{Introduction}
\label{sec:introduction}
Scale is inherent in the structure of the physical world around us and the measurements we make of it. Ideally, the machine learning models we run on this perceptual data should have a notion of scale, which is either learnt or built directly into them. However, the state-of-the-art models of our time, convolutional neural networks (CNNs) \citep{LecunBBH98}, are predominantly local in nature due to small filter sizes. It is not thoroughly understood how they account for and reason about multiscale interactions in their deeper layers, and empirical evidence \citep{ChenPKMY18,YuK15,YuKF17} using dilated convolutions suggests that there is still work to be done in this arena.

In computer vision, typical methods to circumvent scale are: \emph{scale averaging}, where multiple scaled versions of an image are fed through a network and then averaged \citep{Kokkinos15}; \emph{scale selection}, where an object's scale is found and then the region is resized to a canonical size before being passed to a network \citep{GirshickDDM14}; and \emph{scale augmentation}, where multiple scaled versions of an image are added to the training set \citep{BarnardC91}. While these methods help, they lack explicit mechanisms to fuse information from different scales into the same representation. In this work, we construct a generalized convolution taking, as input, information from different scales.

The utility of convolutions arises in scenarios where there is a \emph{translational symmetry} (translation invariance) inherent in the task of interest \citep{CohenW16}. Examples of such tasks are object classification \citep{KrizhevskySH12}, object detection \citep{GirshickDDM14}, or dense image labelling \citep{LongSD15}. By using translational weight-sharing \citep{LecunBBH98} for these tasks, we reduce the parameter count while preserving symmetry in the deeper layers. The overall effect is to improve sample complexity and thus reduce generalization error \citep{SokolicGSR17}. Furthermore, it has been shown that convolutions (and various reparameterizations of them) are the only linear operators that preserve symmetry \citep{KondorT18}. Attempts have been made to extend convolutions to scale, but they either suffer from breaking translation symmetry \citep{HenriquesV17, EstevesAZD17}, making the assumption that scalings can be modelled in the same way as rotations \citep{MarcosKLT18}, or ignoring symmetry constraints \citep{Hilbert18}. The problem with the aforementioned approaches is that they fail to account for the unidirectional nature of scalings. In data there exist many one-way transformations, which cannot be inverted. Examples are occlusions, causal translations, downscalings of discretized images, and pixel lighting normalization. In each example the transformation deletes information from the original signal, which cannot be regained, and thus it is non-invertible. We extend convolutions to these classes of symmetry under noninvertible transformations via the theory of semigroups. Our contributions are the introduction of a semigroup equivariant correlation and a scale-equivariant CNN.

\begin{figure*}
    \centering
    \includegraphics[width=\linewidth]{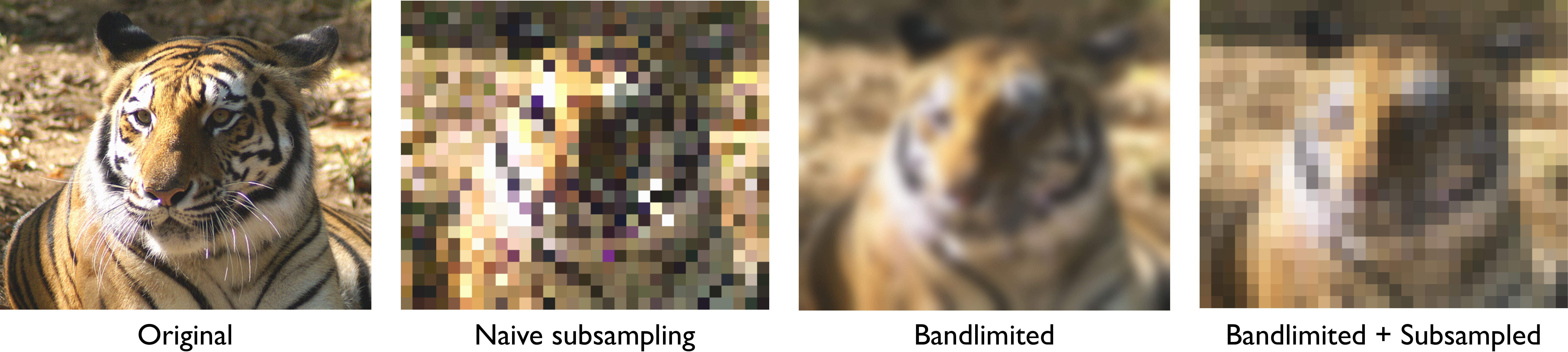}
    \caption{How to correctly downsample an image. Left to right: The original high-resolution image; an $\times1/8$ subsampled image, notice how a lot of the image structure has been destroyed; a high-resolution, bandlimited (blurred) image; a bandlimited and $\times1/8$ subsampled image. Compare the bandlimited and subsampled image with the na\"ively subsampled image. Much of the low-frequency image structure is preserved in the bandlimited and subsampled image. Image source: ImageNet.}
    \label{fig:downsampling}
\end{figure*}

\section{Background}
This section introduces some key concepts such as groups, semigroups, actions, equivariance, group convolution, and scale-spaces. These concepts are presented for the benefit of the reader, who is not expected to have a deep knowledge of any of these topics \emph{a priori}. 

\textbf{Downsizing images}
We consider sampled images $f \in L^2(\mbb{Z}^2)$ such as in Figure \ref{fig:downsampling}. For image $f$, $x$ is pixel position and $f(x)$ is pixel intensity. If we wish to downsize by a factor of 8, a na\"ive approach would be to subsample every 8\textsuperscript{th} pixel: $f_\text{down}(x) = f(8x)$. This leads to an artifact, \emph{aliasing} \citep[p.43]{Mallat09}, where the subsampled image contains information at a higher-frequency than can be represented by its resolution. The fix is to \emph{bandlimit} pre-subsampling, suppressing high-frequencies with a blur. Thus a better model for downsampling is $f_\text{down}(x) = [G *_{\mbb{Z}^d} f](8x)$, where $*_{\mbb{Z}^d}$ denotes convolution over $\mbb{Z}^d$, and $G$ is an appropriate blur kernel (discussed later). Downsizing involves necessary information loss and cannot be inverted \citep{Lindeberg97}. Thus upsampling of images is not well-defined, since it involves imputing high-frequency information, not present in the low resolution image. As such, \emph{in this paper we only consider image downscaling}.

\begin{wrapfigure}{r}{0.28\linewidth}
	\centering
    \includegraphics[width=\linewidth]{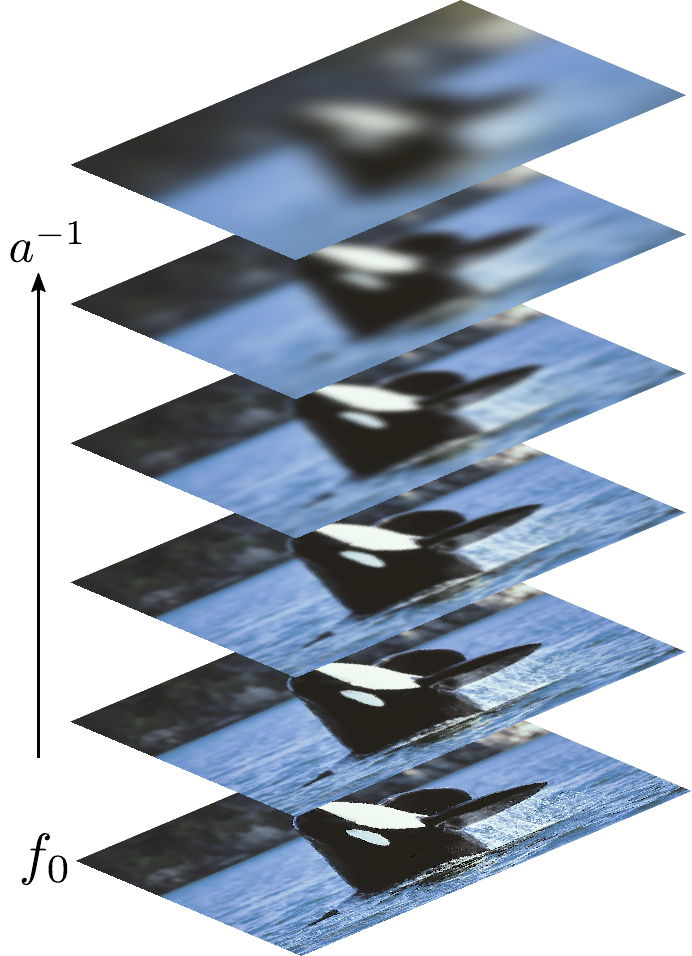}
    \caption{A Scale-space: For implementations we logarithmically discretize the scale-axis in $a^{-1}$.}
    \label{fig:scalespace}
\end{wrapfigure}

\textbf{Scale-spaces}
Scale-spaces have a long history, dating back to the late fifties with the work of \citet{iijima59}. They consist of an image $f_0 \in L^2(\mbb{R}^d)$ and multiple blurred versions of it. Although sampled images live on $\mbb{Z}^d$, scale-space analysis tends to be over $\mbb{R}^d$, but many of the results we present are valid on both domains. Among all variants, the Gaussian scale-space (GSS) is the commonest \citep{Witkin83}. Given an initial image $f_0$, we construct a GSS by convolving $f_0$ with an isotropic (rotation invariant) Gauss-Weierstrass kernel $G(x,t) = (4\pi t)^{-d/2} \exp\left\{ \|x\|^2 / 4t \right\}$ of variable width $\sqrt{t}$ and spatial positions $x$. The GSS is the complete set of responses $f(t,x)$:
\begin{align}
    f(t,x) &= [G(\cdot,t) *_{\mbb{R}^d} f_0](x), \qquad t > 0 \\
    f(0,x) &= f_0(x),
\end{align}
where $*_{\mbb{R}^d}$ denotes convolution over $\mbb{R}^d$. The higher the level $t$ (larger blur) in the image, the more high frequency details are removed. An example of a scale-space $f(t,x)$ can be seen in Figure \ref{fig:scalespace}. An interesting property of scale-spaces is the \emph{semigroup property} \citep{FlorackRKV92}, sometimes referred to as the \emph{recursivity principle} \citep{PauwelsGFM95}, which is 
\begin{align}
    f(s+t, \cdot) = G(\cdot, s) *_{\mbb{R}^d} f(t, \cdot) \label{eq:semigroup_property}
\end{align}
for $s, t > 0$. It says that we can generate a scale-space from other levels of the scale-space, not just from $f_0$. Furthermore, since $s,t >0$ it also says that we cannot generate sharper images from blurry ones, using just a Gaussian convolution. Thus moving to blurrier levels encodes a degree of information loss. This property emanates from the closure of Gaussians under convolution, namely for multidimensional Gaussians with covariance matrices $\Sigma$ and $T$
\begin{align}
    G(\cdot, \Sigma + T) = G(\cdot, \Sigma) *_{\mbb{R}^d} G(\cdot, T).
\end{align}
We assume the initial image $f_0$ has a maximum spatial frequency content---dictated by pixel pitch in discretized images---which we model by assuming the image has already been convolved with a width $s_0$ Gaussian, which we call the \emph{zero-scale}. Thus an image of bandlimit $s$ in the scale-space is found at GSS slice $f(s-s_0,\cdot)$, which we see from Equation \ref{eq:semigroup_property}. There are many varieties of scale-spaces: the $\alpha$-scale-spaces \citep{PauwelsGFM95}, the discrete Gaussian scale-spaces \citep{Lindeberg90}, the binomial scale-spaces \citep{Burt81}, etc. These all have specific kernels, analogous to the Gaussian, which are closed under convolution (details in supplement).

Slices at level $t$ in the GSS correspond to images downsized by a \emph{dilation factor} $0\leq a \leq 1$ ($1=\text{no downsizing}$, $0=\text{shrinkage to a point}$). An $a$-dilated and appropriately bandlimited image $p(a,x)$ is found as (details in supplement)
\begin{align}
    p(a,x) = f(t(a,s_0), a^{-1}x), \qquad t(a,s_0) := \frac{s_0}{a^2} - s_0.
\end{align}
For clarity, we refer to decreases in the spatial dimensions of an image as \emph{dilation} and increases in the blurriness of an image as \emph{scaling}. For a generalization to anisotropic scaling we replace scalar scale $t$ with matrix $T$, zero-scale $s_0$ with covariance matrix $\Sigma_0$, and dilation parameter $a$ with matrix $A$, so
\begin{align}
    T(A,\Sigma_0) = A^{-1} \Sigma_0 A^{-T} - \Sigma_0. \label{eq:anisotropic}
\end{align}
\textbf{Semigroups}
The semigroup property of Equation \ref{eq:semigroup_property} is the gateway between classical scale-spaces and the group convolution \citep{CohenW16} (see end of section). Semigroups $(S,\circ)$ consist of a non-empty set $S$ and a (binary) composition operator $\circ: S \times S \to S$. Typically, the composition $s\circ t$ of elements $s, t \in S$ is abbreviated $st$. For our purposes, these individual elements will represent dilation parameters. For $S$ to be a semigroup, it must satisfy the following two properties
\begin{itemize}
    \item Closure: $st \in S$ for all $s,t \in S$
    \item Associativity: $(st)r = s(tr) = str$ for all $s,t,r\in S$
\end{itemize}
Note that commutativity is not a given, so $st \neq ts$. The family of Gaussian densities under spatial convolution is a semigroup\footnote{For the Gaussians we find the identity element as the limit $\lim_{t\downarrow 0} G(\cdot,t)$, which is a Dirac delta. Note that this element is not strictly in the set of Gaussians, thus the Gaussian family has no identity element.}. Semigroups are a generalization of groups, which are used in \citet{CohenW16} to model invertible transformations. For a semigroup to be a group, it must also satisfy the following conditions 
\begin{itemize}
    \item Identity element: there exists an $e\in S$ such that $es=se=s$ for all $s\in S$
    \item Inverses: for each $s\in S$ there exists a $s^{-1}$ such that $s^{-1}s=ss^{-1}=e$.
\end{itemize}

\textbf{Actions}
Semigroups are useful because we can use them to model transformations, also known as (semigroup) \emph{actions}. Given a semigroup $S$, with elements $s\in S$ and a domain $X$, the action $L_s: X \to X$ is a map, also written $x \mapsto L_s[x]$ for $x\in X$. The defining property of the semigroup action (Equation \ref{eq:left_action}) is that it is associative and closed under composition, essentially inheriting its compositional structure from the semigroup $S$. This action is in fact called a \emph{left} action\footnote{A \emph{right} action is $R_{st}[x] = R_t [R_s [x]]$. Notice $R_{st}$ applies $R_s$ then $R_t$, opposite in order to the left action.}. Notice that for a composed action $L_{st}$, we first apply $L_t$ then $L_s$. 
\begin{align}
    L_{st} [x] = L_s [L_t [x]]. \label{eq:left_action}
\end{align}
Actions can also be applied to functions $f: X \to Y$ by viewing $f$ as a point in a function space. The result of the action is a new function, denoted $L_s[f]$. Since the domain of $f$ is $X$, we commonly write $L_s[f](x)$. Say the domain is a semigroup $S$ then an example left action\footnote{This is indeed a left action because $L_s[L_t[f]](x) = L_t[f](xs) = f(xst) = L_{st}[f](x)$.} is $L_s[f](x) = f(xs)$. Another example, which we shall use later, is the action $S_{A,z}^{\Sigma_0}$ used to form scale-spaces, namely
\begin{align}
    S_{A,z}^{\Sigma_0}[f_0](x) &= [G_A^{\Sigma_0} *_{\mbb{Z}^d} f_0](A^{-1}x+z), \qquad G_A^{\Sigma_0} := G(\cdot, A^{-1}\Sigma_0A^{-\top} - \Sigma_0). \label{eq:scale-space-action}
\end{align}
The elements of the semigroup are the tuples $(A,z)$ (dilation $A$, shift $z$) and $G_A^{\Sigma_0}$ is an anisotropic discrete Gaussian. The action first bandlimits by $G_A^{\Sigma_0} *_{\mbb{Z}^d} f_0$ then dilates and shifts the domain by $A^{-1}x+z$. Note for fixed $(A,z)$ this maps functions on $\mbb{Z}^d$ to functions on $\mbb{Z}^d$.

\textbf{Lifting}
We can also view actions from functions on $X$ to functions $X$ as maps from functions on $X$ to functions on the semigroup $S$. We call this a \emph{lift} \citep{KondorT18}, denoting lifted functions as $f^\uparrow$. One example is the scale-space action (Equation \ref{eq:scale-space-action}). If we set $x=0$, then 
\begin{align}
    f^\uparrow(A,z) = S_{A,z}^{\Sigma_0}[f_0](0) = [G_A^{\Sigma_0} *_{\mbb{R}^d} f_0](z), \label{eq:lifting}
\end{align}
which is the expression for an anisotropic scale-space (parameterized by the dilation $A$ rather than the scale $T(A,\Sigma_0) = A^{-1} \Sigma_0 A^{-T} - \Sigma_0$). To lift a function on to a semigroup, we do not necessarily have to set $x$ equal to a constant, but we could also integrate it out. An important property of \emph{lifted} functions is that actions become simpler. For instance, if we define $f^\uparrow(s) = L_s[f](0)$ then
\begin{align}
    (L_t[f])^\uparrow(s) = L_s[L_t[f]](0) = L_{st}[f](0) = f^{\uparrow}(st). \label{eq:lifted_equivariance}
\end{align}
The action on $f$ could be complicated, like a Gaussian blur, but the action on $f^\uparrow$ is simply a `shift' on $f^{\uparrow}(s) \mapsto f^{\uparrow}(st)$. We can then define the action $L_t$ on lifted functions as $L_t[f^\uparrow](s) = f^{\uparrow}(st)$. 

\textbf{Equivariance and Group Correlations}
\label{sec:group_convolution}
CNNs rely heavily on the (cross-)correlation\footnote{In the deep learning literature these are inconveniently referred to as convolutions, but we stick to correlation.}. Correlations $\star_{\mbb{Z}^d}$ are a special class of linear map of the form
\begin{align}
    [f \star_{\mbb{Z}^d} \psi](s) = \sum_{x\in\mbb{Z}^d} f(x) \psi(x-s).
\end{align}
Given a signal $f$ and filter $\psi$, we interpret the correlation as the collection of inner products of $f$ with all $s$-translated versions of $\psi$. This basic correlation has been extended to transformations other than translation via the \emph{group correlation}, which as presented in \citet{CohenW16} are
\begin{align}
    [f \star_{H} \psi](s) = \sum_{x\in H} f(x) \psi(L_{s^{-1}}[x]),
\end{align}
where $H$ is the relevant group and $L_s$ is a \emph{group} action \eg for rotation $L_{s}[x] = R_s x$, where $R_s$ is a rotation matrix. Note how the summation domain is $H$, as are the domains of the signal and filter. Most importantly, the domain of the output is $H$. It truly highlights how this is an inner product of $f$ and $\psi$ under all $s$-transformations of the filter. The correlation exhibits a very special property. It is \emph{equivariant} under actions of the group $H$. This means
\begin{align}
    L_s [f \star_{H} \psi] = L_s [f] \star_{H} \psi. \label{eq:equivariance}
\end{align}
That is $L_s$ and $\star_H$ commute---group correlation followed by the action is equivalent to the action followed by the group correlation. Note, the group action may `look' different depending on whether it was applied before or after the group correlation, but it \emph{represents} the exact same transformation. 

\section{Method}
We aim to construct a scale-equivariant convolution. We shall achieve this here by introducing an extension of the correlation to semigroups, which we then tailor to scalings.

\textbf{Semigroup Correlation}
\label{sec:semigroup_correlation}
There are multiple candidates for a semigroup correlation $\star_S$. The basic ingredients of such a correlation will be the inner product, the semigroup action $L_s$, and the functions $\psi$ and $f$. Furthermore, it must be equivariant to (left) actions on $f$. For a semigroup $S$, domain $X$, and action $L_s: X\to X$, we define:
\begin{align}
	[\psi \star_S f](s) = \sum_{x\in X} \psi(x) L_s [f](x).
\end{align}
It is the set of responses formed from taking the inner product between a filter $\psi$ and a signal $f$ under all transformations of the signal. Notice that we transform the signal and not the filter and that we write $\psi \star_S f$, not $f \star_S \psi$---it turns out that a similar expression where we apply the action to the filter is not equivariant to actions on the signal. Furthermore this expression \emph{lifts} a function from $X$ to $S$, so we expect actions on $f$ to look like a `shift' on the semigroup. A proof of equivariance to (left) actions on $f$ is as follows
{
\medmuskip=4mu
\thinmuskip=4mu
\thickmuskip=4mu
\begin{align}
	[\psi \star L_t[f]](s) &= \sum_{x\in X} \psi(x) L_s [L_t [f]](x) = \sum_{x\in X} \psi(x) L_{st} [f](x) = [\psi \star f](st) = L_t [\psi \star f](s) \label{eq:s-equivariance}
\end{align}
}\noindent
We have used the definition of the left action $L_sL_t = L_{st}$, the semigroup correlation, and our definition of the action for lifted functions. We can recover the standard `convolution', by substituting $S=\mbb{Z}^d$, $X=\mbb{Z}^d$, and the translation action $L_s[f](x) = f(x+s)$:
\begin{align}
	[\psi \star_{\mbb{Z}^d} f](s) &= \sum_{x\in \mbb{Z}^d} \psi(x) f(x+s) = \sum_{x'\in \mbb{Z}^d} \psi(x'-s) f(x').
\end{align}
where $x'=x+s$. We can also recover the group correlation by setting $S=H$, $X=H$, where $H$ is a discrete group, and $L_s[f](x) = f(L'_{s}x)$, where $L'_s$ is the form of $L_s$ acting on the domain $X=H$:
\begin{align}
	[\psi \star_{H} f](s) &= \sum_{x\in H} \psi(x) f(L'_s[x]) = \sum_{x'\in H} \psi(L'_{s^{-1}}[x']) f(x').
\end{align}
where $x'=L'_s[x]$ and since $L'$ is a \emph{group} action inverses exist, so $x = L'_{s^{-1}}[x']$. The semigroup correlation has a two notable differences from the group correlation: i) In the semigroup correlation we transform the signal and not the filter. When we restrict to the group and standard convolution, transforming the signal or the filter are equivalent operations since we can apply a change of variables. This is not possible in the semigroup case, since this change of variables requires an inverse, which we do not necessarily have. ii) In the semigroup correlation, we apply an action to the \emph{whole} signal as $L_s[f]$, as opposed to just the domain ($f(L_s[x])$). This allows for more general transformations than allowed by the group correlation of \citet{CohenW16}, since transformations of the form $f(L_s[x])$ can only move pixel locations, but transformations of the form $L_s[f]$ can alter the values of the pixels as well, and can incorporate neighbourhood information into the transformation. 

\textbf{The Scale-space Correlation}
We now have the tools to create a scale-equivariant correlation. All we have to choose is an appropriate action for $L_s$. We choose the scale-space action of Equation \ref{eq:scale-space-action}. The scale-space action $S_{z,A}^{\Sigma_0}$ for functions on $\mbb{Z}^d$ is given by
\begin{align}
    S_{A,z}^{\Sigma_0}[f](x) = [G_A^{\Sigma_0} *_{\mbb{Z}^d} f](A^{-1}x+z), \qquad S_{A,z}^{\Sigma_0} S_{B,y}^{\Sigma_0} = S_{AB,Ay+z}^{\Sigma_0}.
\end{align}

\begin{figure*}
    \centering
    \includegraphics[width=\linewidth]{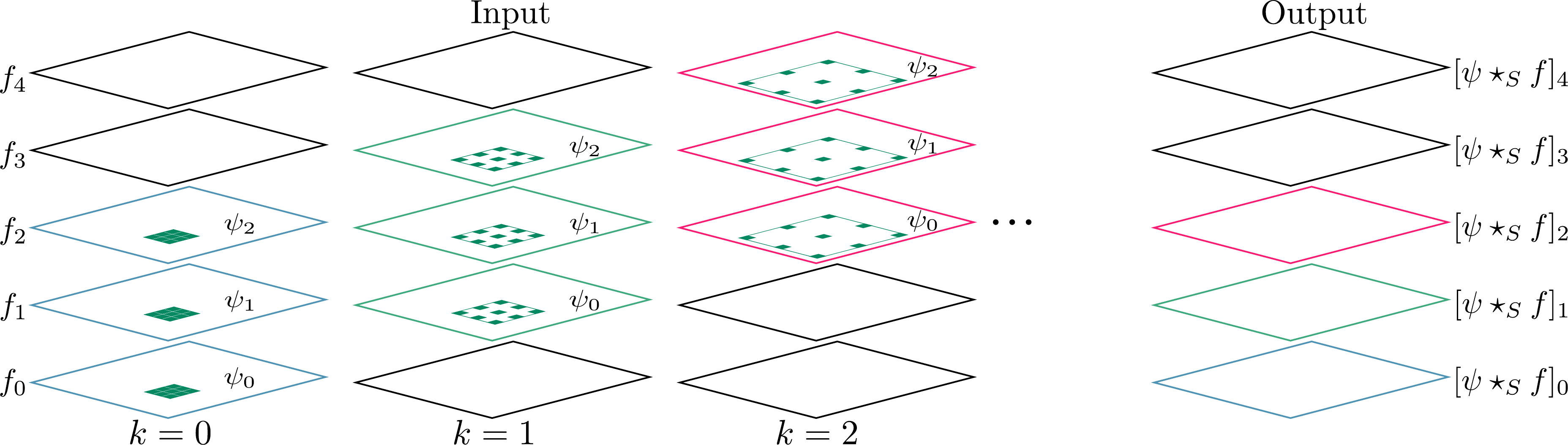}
    \caption{Scale correlation schematic: The left 3 stacks are the same input $f$, with levels $f_\ell(\cdot) = f(2^{-\ell} I, \cdot)$. Each stack shows the inner product between filter $\psi$ (in green) at translation $z$ for dilation $2^{-k} I$ corresponding to the output level $[\psi \star_S f]_k$ on the right with matching color. Notice that as we dilate the filter, we also shift it one level up in the scale-space, accoridng to Equation \ref{eq:dss-correlation}.}
    \label{fig:scale-space-correlation}
\end{figure*}

Since our scale-space correlation only works for discrete semigroups, we have to find a suitable discretization of the dilation parameter $A$. Later on we will choose a discretization of the form $A_k = 2^{-k} I$ for $k \geq 0$, but for now we will just assume that there exists some countable set $\mc{A}$, such that $S=\{(A,z)_{A\in\mc{A}, z\in\mbb{Z}^d}\}$ is a valid discrete semigroup. We begin by assuming we have lifted an input image $f$ on to the scale-space via Equation \ref{eq:lifting}. The lifted signal is indexed by coordinates $A,z$ and so the filters share this domain and are of the form $\psi(A,z)$. The scale-space correlation is then
\begin{align}
    [\psi \star_S f]&(A,z) = \sum_{(B,y)\in S} \psi(B,y) S_{B,y}^{\Sigma_0} [f](A,y) = \sum_{B\in\mc{A}} \sum_{y\in \mbb{Z}^d} \psi(B,y) f(BA,A^{-1}y+z). \label{eq:space-correlation-activations}
\end{align}
For the second equality we recall the action on a lifted signal is governed by Equation \ref{eq:lifted_equivariance}. The appealing aspect of this correlation is that we do not need to convolve with a bandlimiting filter---a potentially expensive operation to perform at every layer of a CNN---since we use signals that have been lifted on to the semigroup. Instead, the action of scaling by $A$ is accomplished by `fetching' a slice $f(BA,\cdot)$ from a blurrier level of the scale-space. Let's restrict the scale correlation to the scale-space where $A_k = 2^{-k}I$ for $k \geq 0$, with zero-scale $\Sigma_0 = \frac{1}{4}I$. Denoting $f(2^{-k}I, \cdot)$ as $f_k(\cdot)$, this can be seen as a dilated convolution \citep{YuK15} between $\psi_\ell$ and slice $f_{\ell+k}$. This form of the scale-space correlation (shown below) we use in our experiments. A diagram of this can be seen in Figure \ref{fig:scale-space-correlation}:
\begin{align}
    	[\psi \star_S f]_k(z) = \sum_{\ell \geq 0} \sum_{y\in \mbb{Z}^d} \psi_{\ell}(y) f_{\ell+k}(2^{k}y+z). \label{eq:dss-correlation}
\end{align}

\textbf{Equivariant Nonlinearities}
\label{sec:nonlinearities}
Not all nonlinearities commute with semigroup actions, but it turns out that pointwise nonlinearities $\nu$ commute with a special subset of actions of the form, 
\begin{align}
    L_s[f](x) = f(L_s'[x]) \label{eq:pointwise}
\end{align}
where $L_s'$ is a representation of $L_s$ acting on the domain of $f$. For these sorts of actions, we cannot alter the values of the $f$, just the locations of the values. If we write function composition as $[\nu \bullet f](x) = \nu(f(x))$, then a proof of equivariance is as follows:
\begin{align}
    [\nu \bullet L_s [f]](x) &= \nu(f(L'_s[x])) = [\nu \bullet f](L'_s[x]) = L_s [\nu \bullet f](x).
\end{align}
Equation \ref{eq:pointwise} may at first glance seem overly restrictive, but it turns out that this is not the case. Recall that for functions lifted on to the semigroup, the action is $L_s[f](x) = f(xs)$. This satisfies Equation \ref{eq:pointwise}, and so we are free to use pointwise nonlinearities.

\textbf{Batch normalization}
For batch normalization, we compute batch statistics over all dimensions of an activation tensor except its channels, as in \citet{CohenW16}.

\textbf{Initialization}
Since our correlations are based on dilated convolutions, we use the initialization scheme presented in \citep{YuK15}. For pairs of input and output channels the center pixel of each filter is set to one and the rest are filled with random noise of standard deviation $10^{-2}$. 

\textbf{Boundary conditions}
In our semigroup correlation, the scale dimension is infinite---this is a problem for practical implementations. Our solution is to truncate the scale-space to finite scale. This breaks global equivariance to actions in $S$, but is locally correct. Boundary effects occur at activations with receptive fields covering the truncation boundary. To mitigate these effects we: i) use filters with a scale-dimension no larger than two, ii) interleave filters with scale-dimension 2 with filters of scale-dimension 1. The scale-dimension 2 filters enable multiscale interactions but they propagate boundary effects; whereas, the scale-dimension 1 kernels have no boundary overlap, but also no multiscale behavior. Interleaving trades off network expressiveness against boundary effects.

\textbf{Scale-space implementation}
We use a 4 layer scale-space and zero-scale $1/4$ with dilations at integer powers of 2, the maximum dilation is 8 and kernel width 33 (4 std. of a discrete Gaussian). We use the discrete Gaussian of \citet{Lindeberg90}. In 1D for scale parameter $t$, this is
\begin{align}
    G(x, t) = e^{-t}I_{|x|}(t)
\end{align}
where $I_x(t)$ is the modified Bessel function of integer order. For speed, we make use of the separability of isotropic kernels. For instance, convolution with a 2D Gaussian can we written as a convolution with 2 identical 1D Gaussians sequentially along the $x$ and then the $y$-axis. For an $N\times N$ input image and $M\times M$ blurring kernel, this reduces computational complexity of the convolution as $O(M^2N^2) \to O(2MN^2)$. With GPU parallelization, this saving is $O(M^2) \to O(M)$, which is especially significant for us since the largest blur kernel we use has $M=33$. 

\textbf{Multi-channel features}
Typically CNNs use multiple channels per activation tensor, which we have not included in our above treatment. In our experiment we include input channels $i$ and output channels $o$, so a correlation layer is
\begin{align}
    [\psi \star_S f]_k^o(z) = \sum_{i} \sum_{\ell \geq 0} \sum_{x\in \mbb{Z}^d} \psi_{\ell}^{i,o}(x) f_{\ell+k}^{i}(2^{k}x+z).
\end{align}

\section{Experiments and Results}
Here we present our results on the Patch Camelyon \citep{VeelingLWCW18} and Cityscapes \citep{CordtsORREBFRS16} datasets. We also visualize the quality of scale-equivariance achieved.

\begin{table}[t]
\centering
\caption{Results on the Patch Camelyon and Cityscapes Dataset. Higher is better}
\label{tab:pcam_cityscapes}
\begin{tabular}{l c c}
    \hline
	PCam Model		        & Accuracy \\
	\Xhline{2\arrayrulewidth}
	DenseNet Baseline       & 87.0 \\
	S-DenseNet (Ours)       & 88.1 \\
	\citep{VeelingLWCW18}   & 89.8 \\
	\hline
\end{tabular}
\quad
\begin{tabular}{l c c}
    \hline
	Cityscapes Model                & mAP \\
	\Xhline{2\arrayrulewidth}
	ResNet, matched parameters      & 45.66 \\
	ResNet, matched channels        & 49.99 \\
	S-ResNet, multiscale (Ours)	    & 63.53 \\
	S-ResNet, no interaction (Ours) & 64.78 \\
	\hline
\end{tabular}
\end{table}

\textbf{Patch Camelyon}
The Patch Camelyon or PCam dataset \citep{VeelingLWCW18} contains 327 680 tiles from two classes, metastatic (tumorous) and non-metastatic tissue. Each tile is a $96 \times 96$ px RGB-crop labelled as metastatic if there is at least one pixel of metastatic tissue in the central $32\times 32$ px region of the tile. We test a 4-scale DenseNet model \citep{HuangLMW17}, `S-DenseNet', on this task (architecture in supplement). We also train a scale non-equivariant DenseNet baseline and the rotation equivariant model of \citep{VeelingLWCW18}. Our training procedure is: 100 epochs SGD, learning rate 0.1 divided by 10 every 40 epochs, momentum 0.9, batch size of 512, split over 4 GPUs. For data augmentation, we follow the procedure of \citep{VeelingLWCW18, LiuGNDKBVTNCHPS17}, using random flips, rotation, and 8 px jitter. For color perturbations we use: brightness delta 64/255, saturation delta 0.25, hue delta 0.04, constrast delta 0.75. The evaluation metric we test on is accuracy. The results in Table \ref{tab:pcam_cityscapes} show that both the scale and rotation equivariant models outperform the baseline.

\textbf{Cityscapes}
The Cityscapes dataset \citep{CordtsORREBFRS16} contains 2975 training images, 500 validation images, and 1525 test images of resolution $2048 \times 1024$ px. The task is semantic segmentation into 19 classes. We train a 4-scale ResNet \cite{HeZRS16}, `S-ResNet', and baseline. We train an equivariant network with and without multiscale interaction layers. We also train two scale non-equivariant models, one with the same number of channels, one with the same number of parameters. Our training procedure is: 100 epochs Adam, learning rate $10^{-3}$ divided by 10 every 40 epochs, batch size 8, split over 4 GPUs. The results are in Table \ref{tab:pcam_cityscapes}. The evaluation metric is mean average precision. We see that our scale-equivariant model outperforms the baselines. We must caution however, that better competing results can be found in the literature. The reason our baseline underperforms compared to the literature is because of the parameter/channel-matching, which have shrunk its size somewhat due to our own resource constraints. On a like-for-like comparison scale-equivariance appears to help.

\textbf{Quality of Equivariance}
\begin{figure*}[b]
    \centering
    \includegraphics[width=0.325\linewidth]{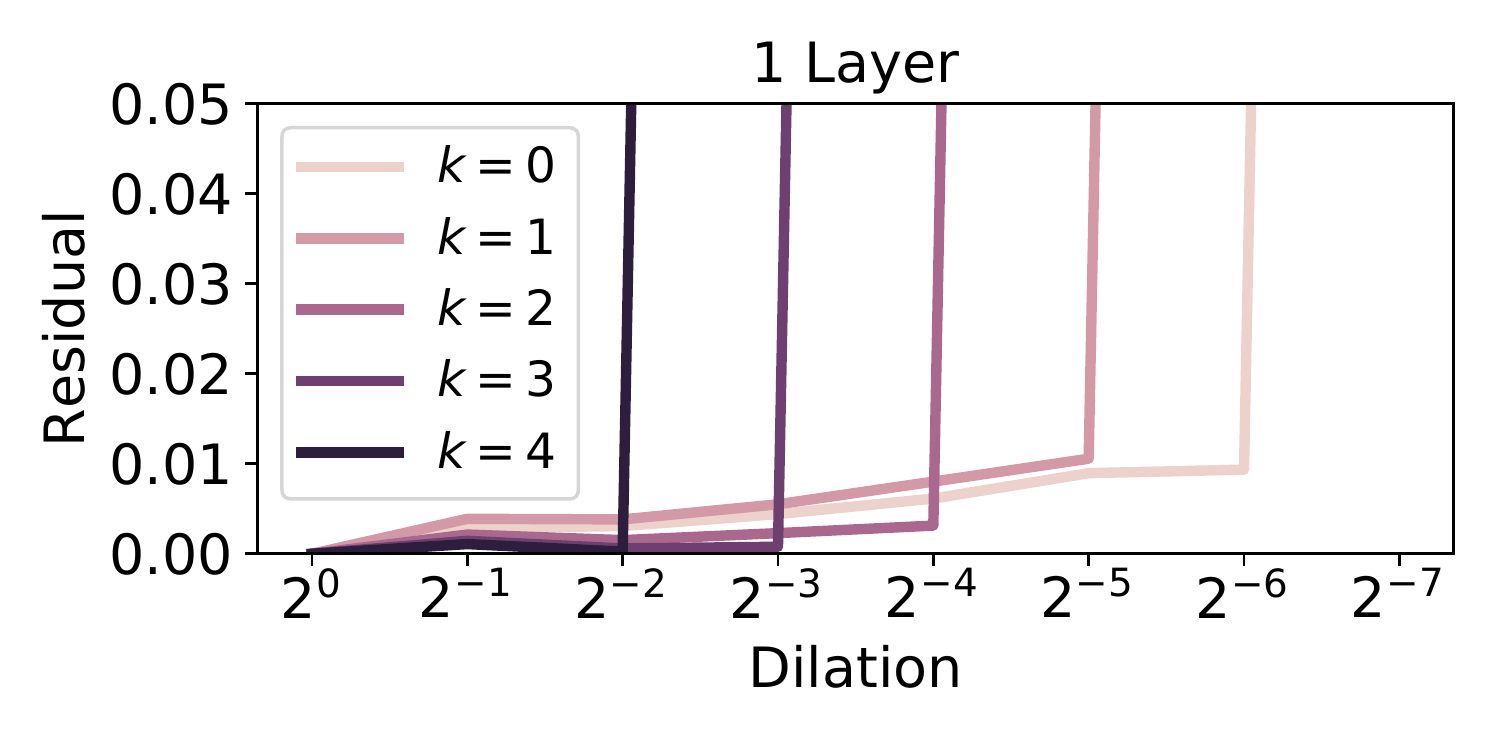}
    \includegraphics[width=0.325\linewidth]{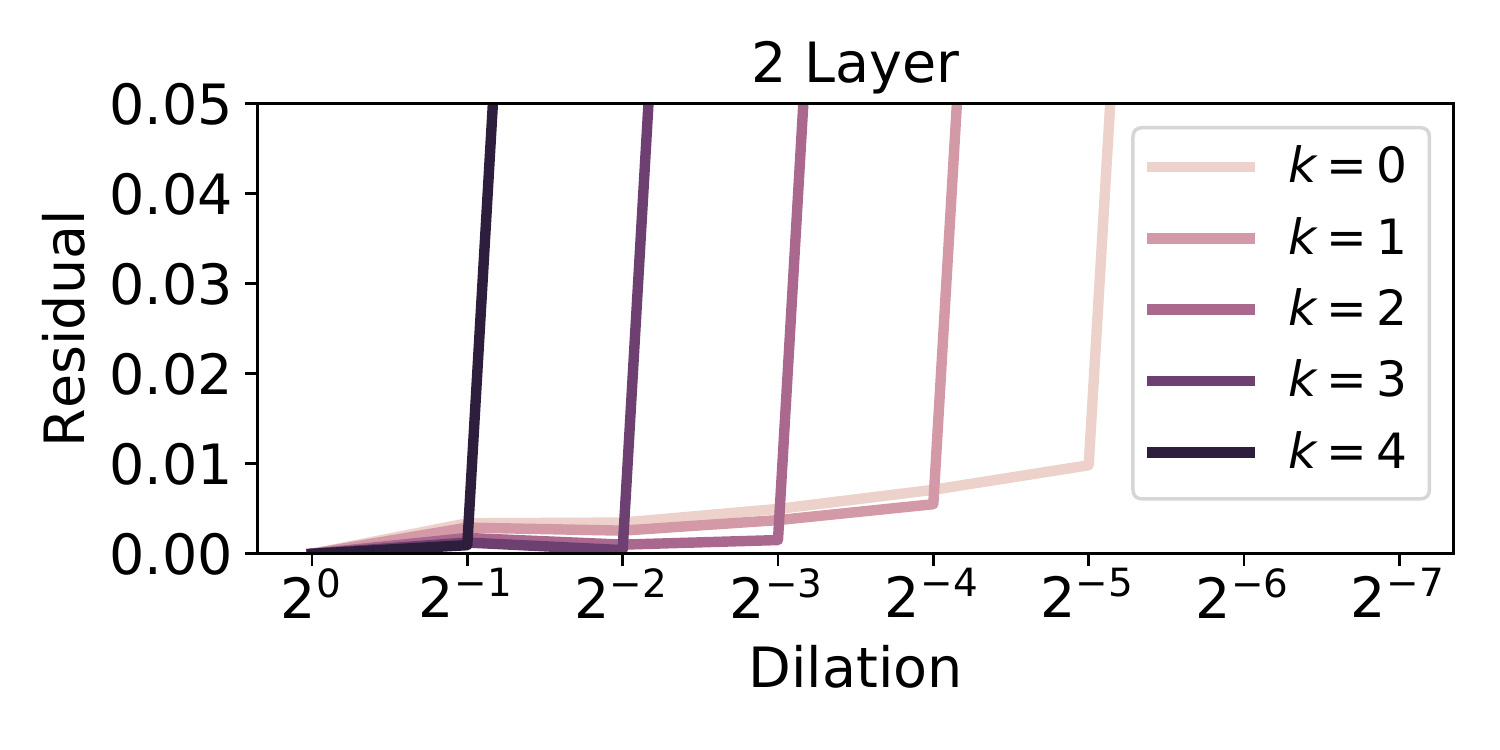}
    \includegraphics[width=0.325\linewidth]{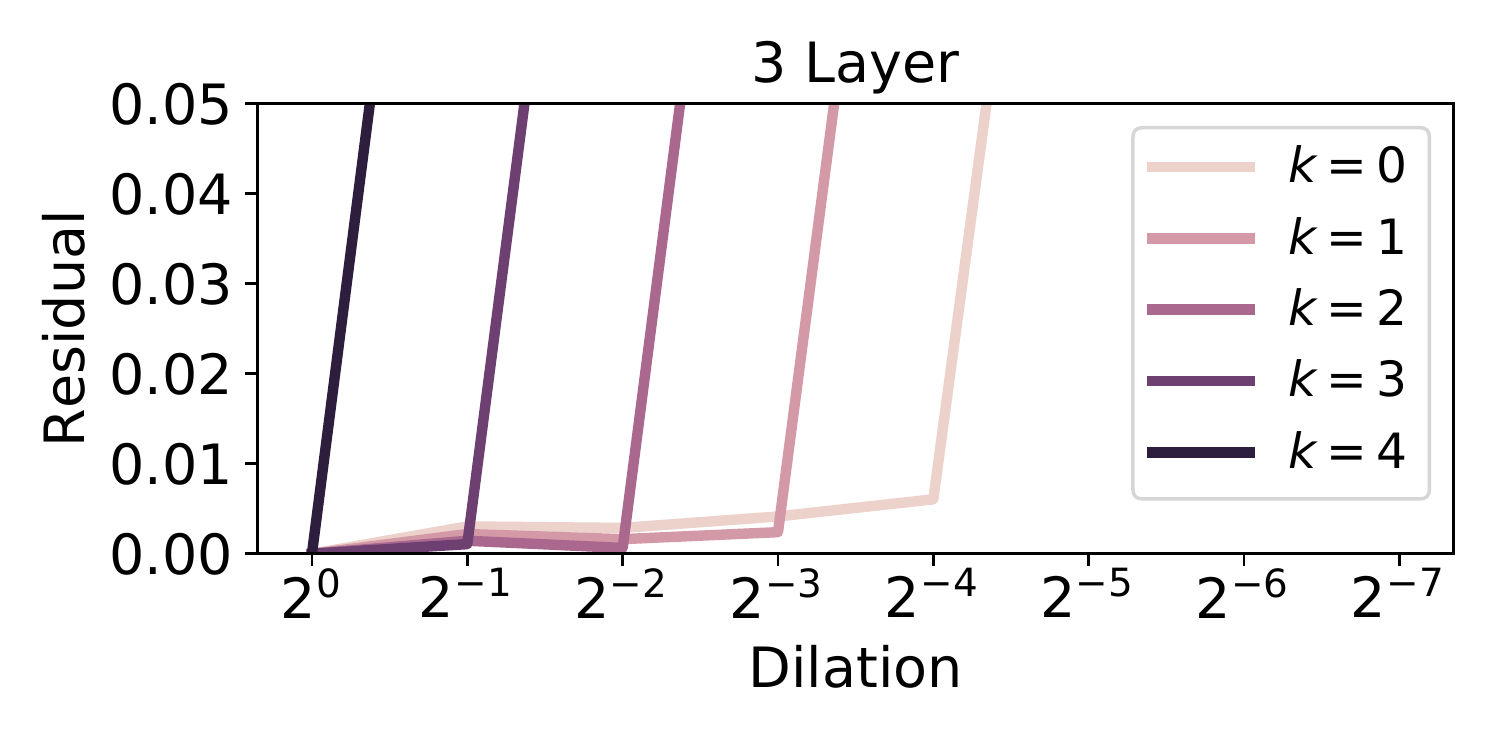}
    \caption{Equivariance quality. Left to right: 1, 2, and 3 layer DSSs. Each line represents the error as in Equation \ref{eq:error}. We see the residual error is typically $<0.01$ until boundary effects are present.}
    \label{fig:activations}
\end{figure*}
We validate the quality of equivariance empirically by comparing activations of a dilated image against the theoretical action on the activations. Using $\Phi$ to denote the deep scale-space (DSS) mapping, we compute the normalized $L2$-distance at each level $k$ of a DSS. Mathematically this is
\begin{align}
    L(2^{-\ell}, k) = \frac{\|\Phi[f](k+\ell,2^{\ell}\cdot) - \Phi [S_{2^{-\ell},0}^{1/4 I}[f]](k,\cdot) \|_2}{\| \Phi[f](k+\ell,2^{\ell}\cdot) \|_2}. \label{eq:error}
\end{align}
The equivariance errors are in Figure \ref{fig:activations} for 3 DSSs with random weights and a scale-space trucated to 8 scales. We see that the average error is below 0.01, indicating that the network is mostly equivariant, with errors due to truncation of the discrete Gaussian kernels used to lift the input to scale-space. We also see that the equivariance errors blow up whenever for constant $\ell + k$ in each graph. This is the point where the receptive field of an activation overlaps with the scale-space truncation boundary.

\section{Related Work}
In recent years, there have been a number of works on group convolutions, namely continuous roto-translation in 2D \citep{WorrallGTB17} and 3D \citep{WeilerGWBC18, KondorLT18, ThomasSKYLKR18} and discrete roto-translations in 2D \citep{CohenW16, WeilerHS18, BekkersLVEPD18, HoogeboomPCW18} and 3D \citep{WorrallGTB17}, continuous rotations on the sphere \citep{EstevesAMD18, CohenGKW18}, in-plane reflections \citep{CohenW16}, and even reverse-conjugate symmetry in DNA sequencing \citep{LunterB17}. Theory for convolutions on compact groups---used to model invertible transformations---also exists \citep{CohenW16, CohenW16a, KondorT18, CohenGW18}, but to date the vast majority of work has focused on rotations. 

For scale, there far fewer works. \citep{HenriquesV17, EstevesAZD17} both perform a log-polar transform of the signal before passing to a standard CNN. Log-polar transforms reparameterize the input plane into angle and log-distance from a predefined origin. The transform is sensitive to origin positioning, which if done poorly breaks translational equivariance. \citep{MarcosKLT18} use a group CNN architecture designed for roto-translation, but instead of rotating filters in the group correlation, they scale them. This seems to work on small tasks, but ignores large scale variations. \citep{Hilbert18} instead use filters of different sizes, but without any attention to equivariance.

\section{Discussion, Limitations, and Future Works}
We found the best performing architectures were composed mainly of correlations where the filters' scale dimension is one, interleaved with correlations where the scale dimension is higher. This is similar to a network-in-network \citep{LinCY13} architecture, where $3\times 3$ convolutional layers are interleaved with $1\times 1$ convolutions. We posit this is the case because of boundary effects, as were observed in Figure \ref{fig:activations}. We suspect working with dilations with smaller jumps in scale between levels of the scale-space would help, so $A_k = c^{-k}I$ for $1<c<2$. This is reminiscent of the tendency in the image processing community to use non-integer scales. This would, however, involve the development of non-integer dilations and hence interpolation. We see working on mitigating boundary effects as an important future work, not just for scale-equivariance, but CNNs as a whole.

Another limitation of the current model is the increase in computational overhead, since we have added an extra dimension to the activations. This may not be a problem long term, as GPUs grow in speed and memory, but the computational complexity of a correlation grows exponentially in the number of symmetries of the model, and so we need more efficient methods to perform correlation, either exactly or approximately.

We see semigroup correlations as an exciting new family of operators to use in deep learning. We have demonstrated a proof-of-concept on scale, but there are many semigroup-structured transformations left to be explored, such as causal shifts, occlusions, and affine transformations. Concerning scale, we are also keen to explore how multi-scale interactions can be applied on other domains as meshes and graphs, where symmetry is less well-defined.

\section{Conclusion}
We have presented deep scale-spaces a generalization of convolutional neural networks, exploiting the scale symmetry structure of conventional image recognition tasks. We outlined new theory for a generalization of the convolution operator on semigroups, the semigroup correlation. Then we showed how to derive the standard convolution and the group convolution \citep{CohenW16} from the semigroup correlation. We then tailored the semigroup correlation to the scale-translation action used in classical scale-space theory and demonstrated how to use this in modern neural architectures.

\newpage
\bibliographystyle{plainnat}
\bibliography{example_paper}

\newpage
\appendix

{\Large\textbf{Supplementary Material}}

In this supplementary material we elaborate on where the scale-space action comes from and the architectures used in our paper.

\section{Scale-spaces}
Here we provide some extra information on scale-spaces for the interested reader. For in depth literature, we suggest \cite{FlorackRKV94a, FlorackRKV92, PauwelsGFM95, Lindeberg90, Lindeberg97, crowley2002fast, SaldenRV98a, DuitsFGR04, DuitsFFP03, DuitsB07, BurgethDW05, BurgethDW05b}.

\subsection{1D Gaussian Scale-space}
We are given an initial 1D signal $f_0$ with intrinsic bandlimit, or \emph{zero-scale}, defined by $s_0$ \ie, we impose a maximum frequency, such that there is a correspondence with discretized signals. Note that $\sqrt{s_0}$ is inversely proportional to frequency content of the signal. We wish to downsize it isotropically by a factor $a$, which we call the \emph{dilation}. For this we introduce the downsizing action $L_a[f](x) = f(a^{-1}x)$ for $a \leq 1$. We model the bandlimit of the initial signal as the result of convolving some other signal $f$ with a Gaussian of width $s_0$, so
\begin{align}
    f_0(x) = [G(\cdot, s_0) *_{\mbb{R}^d} f](x).
\end{align}
Now the result of the downscaling action of $f_0$ is as follows
\begin{align}
    L_a[f_0](x) &= f_0(a^{-1}x) = \int_{\mbb{R}^d} G(a^{-1}x - y, s_0) f(y) \, \dd y = \int_{\mbb{R}^d} G(x - ay, a^2 s_0) f(y) \, a \dd y \\
    &= \int_{\mbb{R}^d} G(x - z, a^2 s_0) f(a^{-1}z) \, \dd z = [G(\cdot, a^2 s_0) *_{\mbb{R}^d} L_a[f]](x).
\end{align}
From the first to second lines we have performed a change of variables $z=ay$. So we see that the effect of downsizing a bandlimited signal by $a$ shifts the bandlimit from $s_0$ to $a^2 s_0$. Since $a \leq 1$, this means the blurring Gaussian is narrower and so the frequency content of the signal has been shifted higher. The key relation to bear in mind is the shift $s_0 \mapsto a^2 s_0$.

For a proper scaling, we want the result of the downscaling to have the \emph{same bandlimit} as the original signal $f_0$. This is because if we are representing the signal on a discrete grid, then we have a physically defined maximum frequency content we can store, given by the pixel separation. The solution is to convolve the signal $L_a[f_0]$ with a correcting Gaussian of width $s_0 - a^2 s_0$. Note that this is possible, since $a \leq 0$, so $s_0 - a^2 s_0 > 0$. Alternatively, we want to find a correcting Gaussian to blur before downsizing. Say this correcting Gaussian has bandlimit $t$, then we have
\begin{align}
    L_a[G(\cdot, t) * f_0] &= L_a[G(\cdot, t + s_0) * f] = [G(\cdot, a^2(t + s_0)) * L_a[f]].
\end{align}
But we also want the bandlimit of the downscaled signal to be $s_0$, so we have the relation
\begin{align}
    a^2(t + s_0) &= s_0. \label{eq:resolution_balance}
\end{align}
Thus a 1D Gaussian scale-space, parameterized by dilation, can be built by setting
\begin{align}
    f(t(a,s_0),x) &= [G(\cdot, t(a,s_0)) * f_0](x), \qquad t(a,s_0) := \frac{s_0}{a^2} - s_0 \\
    f(0,x) &= f_0(x)
\end{align}

\subsection{$N$D Gaussian Scale-space}
We now explore downscaling in $N$-dimensions. We first of all represent an initial image $f_0$ as
\begin{align}
	f_0 = G(\cdot,\Sigma_0) *_{\mbb{R}^d} f.
\end{align}
We use the Gaussian to represent the fact that $f_0$ should have an intrinsic bandlimit, usually defined by the resolution at which it is sampled. Now let's introduce the affine action:
\begin{align}
	L_{A,z}[f](x) = f(A^{-1}(x - z)).
\end{align}
It simply applies an affine transformation to our signal. Now using a similar logic to in the 1D case, if we concatenate the affine action with bandlimiting we get
\begin{align}
	L_{A,z}[G(\cdot,\Sigma_0) *_{\mbb{R}^d} f_0] = G(\cdot,A\Sigma_0A^\top) *_{\mbb{R}^d} L_{A,z}[f_0]. \label{eq:affine_commutativity}
\end{align}
So we see that resizing a bandlimited signal shifts the bandlimit according to $\Sigma_0 \mapsto A\Sigma_0A^\top$. Since we would like to have the same bandlimit on our signal before and after the resizing (since we can only represent the signal at constant resolution), we introduce a second bandlimiting by convolving with a Gaussian of width $\Sigma_0 - A\Sigma_0A^\top$. To save space, we write $G_\Sigma = G(\cdot,\Sigma)$. So
\begin{align}
	G_{\Sigma_0} *_{\mbb{R}^d} L_{A,z}[f] &= G_{\Sigma_0 - A\Sigma_0A^\top} *_{\mbb{R}^d} L_{A,z}[f_0] = L_{A,z}[G_{A^{-1}\Sigma_0A^{-\top} - \Sigma_0} *_{\mbb{R}^d} f_0]
\end{align}
From the first to the second equality, we have exchanged the order of the Gaussian convolution and the affine action and altered the bandwidth from $\Sigma_0 - A\Sigma_0A^\top$ to $A^{-1}\Sigma_0A^{-\top} - \Sigma_0$, which comes from the relation established in Equation \ref{eq:affine_commutativity}. Thus the affine action for an affine scale-pyramid is defined as
\begin{align}
	L_{A,z}[G_{A^{-1}\Sigma_0A^{-\top} - \Sigma_0} *_{\mbb{R}^d} f](x) = [G_{A}^{\Sigma_0} *_{\mbb{R}^d} f ](A^{-1}(x - z))
\end{align}
where we have defined $G_{A}^{\Sigma_0} = G_{A^{-1}\Sigma_0A^{-\top} - \Sigma_0}$.

For what values of $A$ and $z$ is this action valid? Let's first focus on $A$. To maintain the zero-scale of $\Sigma_0$, we had to convolve with a Gaussian of width $\Delta = \Sigma_0 - A\Sigma_0A^\top$. Now we know that covariance matrices have to be symmetric $\Delta = \Delta^\top$ and positive definite $\Delta \succ 0$. We see already that it is symmetric, but it is not necessary positive definite. If the base bandlimit is of the form $\Sigma_0 = \sigma_0^2I$ (the original image is isotropically bandlimited), then we can rearrange to
\begin{align}
	\Delta &= \Sigma_0 - A\Sigma_0A^\top = (I - AA^\top) \Sigma_0 \succ 0
\end{align}
This expression is only positive definite if $I - AA^\top \succ 0$; that is
\begin{align}
	I \succ AA^\top.
\end{align}
This condition implies that $A^\top$ is a contraction because
\begin{align}
    v^\top (I - AA^\top) v &= \|v\|_2^2 - \|A^\top z\|_2^2 \geq 0 \implies \|v\|_2^2 &\geq \|A^\top v\|_2^2, \qquad \forall v \in \mbb{R}^N.
\end{align}
Another way of phrasing this is the that the singular values of $A$ may not exceed unity. Note that rotations do not break this constraint. So we see this this model naturally aligns with our notion that we can only model image downscalings, and that upscalings are prohibited.

\subsection{Other Scale-space variants}
We have presented the Gaussian scale-space in $N$D, but there also exists a zoo of other scales-spaces. The most prominent are: the $\alpha$-scale-spaces \citep{PauwelsGFM95}, the discrete Gaussian scale-spaces \citep{Lindeberg90}, and the binomial scale-spaces \citep{Burt81}. In the following, we give a brief introduction to each, exhibited in 1D.

\textbf{$\alpha$-scale-spaces}
$\alpha$-scale-spaces \cite{PauwelsGFM95} are a generalization of the Gaussian space-space in the continuous domain. They are easiest to understand by considering their form in Fourier-space. We begin by considering the Fourier transform of the Gaussian space-space over the spatial dimension
\begin{align}
    \hat{f}(t, \omega) &= \hat{G}(\omega, t) \cdot \hat{f}_0(\omega) \\
    \hat{f}(0, \omega) &= \hat{f}_0(\omega),
\end{align}
where $\hat{f}$ is the Fourier transform of $f$. We are interested in finding a collection of filters like $G$, closed under convolution. In the Fourier domain this corresponds to finding a collection of filters, like $\hat{G}$ closed under multiplication. The Fourier transform of the Gauss-Weierstrass kernel is 
\begin{align}
    G(x, t) = \frac{1}{(4 \pi t)^{1/2}} \exp \left \{ -\frac{x^2}{4t} \right \} \overset{\text{FT}}{\iff} \hat{G}(\omega, t) = \exp \left \{ - \omega^2 t \right \}.
\end{align}
The collection $\{\hat{G}(\omega, t)\}_{t> 0}$ is indeed closed under multiplication and forms a semigroup. To form the $\alpha$-scale-spaces we notice that the Fourier kernel
\begin{align}
    \hat{G}^\alpha(\omega, t) = \exp \left \{ - \omega^{2\alpha} t \right \}
\end{align}
is also closed under multiplication and defines a semigroup. The range of $\alpha$ is typically taken to be $(0,1]$, to make sure that higher levels in the $\alpha$-scale-space are blurrier. Notice that for $\alpha=1$ we return to the standard Gaussian scale-space.

\textbf{Binomial Scale-space}
The binomial scale-space \cite{crowley2002fast} is a discrete scale-space in both the spatial and scale dimensions. It is generated by convolution in $\mbb{Z}$ with the binomial kernel
\begin{align}
    B(x, N) = \left . {}^NC_x \middle / \sum_{x=0}^N {}^NC_x \right . ,\qquad {}^NC_x = \frac{N!}{(N-x)!N!},
\end{align}
where $N > 0$ is the width of the kernel and $0 \leq x \leq N$ is the spatial location of the filter tap. Thus the scale-space is
\begin{align}
    f(N, x) = [B(\cdot, N) *_{\mbb{Z}} f_0](x) \qquad N > 0.
\end{align}
As $N$ grows $B(N, x)$ rapidly converges to a Gaussian kernel of variance $\sigma^2 = N / 4$. The Binomial filters are closed under convolution obeying the semigroup property
\begin{align}
    [B(\cdot, N) *_{\mbb{Z}} B(\cdot, M)](x) = B(x, N+M-1).
\end{align}

\textbf{Discrete Gaussian Scale-space}
The discrete Gaussian scale-space \cite{Lindeberg90} is discrete in the spatial dimension but continuous in the scale dimension, which makes it popular to work with in many practical scale-spaces with non-integer dilation. The scale-space is generated by convolution in $\mbb{Z}$ with the discrete Gaussian kernel
\begin{align}
    G(x, t) = e^{-t}I_{|x|}(t), \qquad I_x(t) = \sum_{m=0}^\infty \frac{\left ( \frac{1}{2} x \right)^{2m + \alpha}}{m! \Gamma(m + \alpha + 1)}
\end{align}
where the term $I_{k}(t)$ is a modified Bessel function of the first kind. These can be implemented easily using \texttt{scipy.special.ive}. The scale-space is formed in the usual way as
\begin{align}
    f(t, x) &= [G(\cdot, t) *_{\mbb{Z}} f_0](x) \qquad t > 0 \\
    f(0, x) &= f_0(x).
\end{align}

\section{Architectures In The Experiments}
In the experiments, we use a DenseNet \cite{HuangLMW17} and a ResNet \cite{HeZRS16}. The architectures are as follows. For the scale equivariant versions, we use 4 scales of a discrete Gaussian scale-space \cite{Lindeberg90}.

\textbf{ResNet}
The residual network consists of a concatenation of residual blocks. A single residual block implements the following 
\begin{align}
    y = x + \mc{F}(x)
\end{align}
where on the RHS we refer to $x$ as the skip connection and $\mc{F}(x)$ as the residual connection. If $x$ has fewer channels than $\mc{F}(x)$, then we pad the missing dimensions with zeros. Each residual connection uses a concatenation of two scale-equivariant correlation interleaved with batch normalization (BN) and a ReLU (ReLU) nonlinearity. These are composed as follows (input left, output right)
\begin{align}
    \text{corr[$1,3,3$] - BN - ReLU - corr[$k,3,3$] - BN}.
\end{align}
where $\text{corr[$k,h,w$]}$ refers to a scale correlation with kernel size $[k,h,w]$ and where $k$ is the number of scale channels, $h$ is the spatial height of the filter, and $w$ is its spatial width. We denote the entire residual block as $\text{res[$k,h,w$]}$. 

\begin{table}[t]
    \centering
    \caption{A residual network. Input at the top. A horizonal line denotes spatial average pooling of stride 2, kernel size 2. Shape is displayed as [scale-space levels, height, width, channels out]. The no-res block denotes a residual block without the skip connection, \ie $y=\mc{F}(x)$. Scale pooling denotes an averaging over all scale dimensions.}
    \label{tab:resnet}
    \begin{tabular}{l  l}
        \hline
        Layer type  & Shape \\
        \Xhline{2\arrayrulewidth}
        res[$k,3,3$]        & $[S, 992, 992, N]$\\
        \hline
        res[$k,3,3$]        & $[S, 496, 496, 2N]$\\
        \hline
        res[$1,3,3$]        & $[S, 248, 248, 4N]$\\
        res[$k,3,3$]        & $[S, 248, 248, 4N]$\\
        \hline
        res[$1,3,3$]        & $[S, 124, 124, 8N]$\\
        res[$1,3,3$]        & $[S, 124, 124, 8N]$\\
        res[$1,3,3$]        & $[S, 124, 124, 8N]$\\
        res[$k,3,3$]        & $[S, 124, 124, 8N]$\\
        no-res[$k,3,3$]     & $[S, 124, 124, 8N]$\\
        scale-pool          & $[1, 124, 124, 8N]$\\
        corr[1,1,1],        & $[1, 124, 124, 19]$\\
        bilinear upsample   & $[1, 992, 992, 19]$ \\
        \hline
    \end{tabular}
\end{table}
\begin{table}[b]
    \centering
    \caption{We match model settings with their names from the paper. Settings are displayed as $[k,S,N]$ or [kernels scale dim., num scales, number of channels].}
    \label{tab:resnet-models}
    \begin{tabular}{l  l}
        \hline
        Model & Settings \\
        \Xhline{2\arrayrulewidth}
        S-ResNet, multiscale interaction    & $[2, 4, 16]$ \\
        S-ResNet no interaction             & $[1, 4, 16]$ \\
        ResNet, matched channels            & $[1, 1, 16]$ \\
        ResNet, matched parameters          & $[1, 1, 18]$ \\
        \hline
    \end{tabular}
\end{table}

The model we use is given in Table \ref{tab:resnet}. It follows the practice of \citet{YuKF17}, who use a bilinear upsampling at the end of the network, since segmentations do not tend to contain high frequency details. In our experiments we use the models shown in Table \ref{tab:resnet-models}

\textbf{DenseNet}
The Dense network \cite{HuangLMW17} consists of a concatenation of 3 dense blocks. Each dense block is composed of layers of the form
\begin{align}
    y_{N+1} = \mc{H} \left ( [y_1, y_2, ..., y_N] \right )
\end{align}
where $[y_1, y_2, ..., y_N]$ is the concatenation of all the previous layers' outputs. Each layer $\mc{H}$ is the composition (input left, output right)
\begin{align}
    \text{BN - ReLU - corr[$k,3,3$]}
\end{align}
where $\text{corr[$k,3,3$]}$ was described in the previous section. We use the notation $\text{dense}_C[k,h,w] \times N$ to denote a dense block with $N$ layers and $C$ output channels per layer. The number of channel outputs remains constant per layer within a dense block. Between dense blocks, we insert transition layer which have the form
\begin{align}
    \text{dense[$1,1,1$]$\times 1$ - pool - dense[$k,h,w$]$\times 1$}.
\end{align}
Here we use a $1\times 1$ convolution to halve the number of output channels, and then perform a spatial average pooling with kernel size 2 and stride 2, followed by a second dense layer. We denote these as $\text{transition}[k,h,w]$. We also use long skip connection between transistion layers. The network we use is shown in Table \ref{tab:densenet}.

\begin{table}[t]
    \centering
    \caption{A DenseNet. Input at the top. For shape, we show the number of scales, the height, the width, and the number of channels. $S$ denotes the number of scales used per layer.}
    \label{tab:densenet}
    \begin{tabular}{l  l}
        \hline
        Layer type  & Shape \\
        \Xhline{2\arrayrulewidth}
        $\text{dense}_{12}[1,3,3]$ $\times 3$   & $[S, 96, 96, 39]$\\
        transition[$3,3,3$]                     & $[S, 48, 48, 19]$ \\
        $\text{dense}_{24}[1,3,3]$ $\times 3$   & $[S, 48, 48, 94]$\\
        transition[$3,3,3$]                     & $[S, 24, 24, 47]$ \\
        $\text{dense}_{48}[1,3,3]$ $\times 3$   & $[S, 24, 24, 213]$\\
        Global average pooling                  & $[1,1,213]$ \\
        Linear layer                            & $[1,1,2]$ \\
        \hline
    \end{tabular}
\end{table}

\end{document}